\documentclass[10pt,twocolumn,letterpaper]{article}

\usepackage[pagenumbers]{iccv} % To force page numbers, e.g. for an arXiv version

\usepackage{multirow}
\usepackage{booktabs}
\usepackage{graphicx}
\usepackage{subcaption}
\usepackage{tabularx}
\usepackage{algorithm}
\usepackage{algorithmic}
\usepackage{makecell}
\usepackage{diagbox}
\usepackage{stfloats}

\newcommand{\blue}[1]{\textbf{\textcolor{blue}{#1}}}
\newcommand{\red}[1]{\textbf{\textcolor{red}{#1}}}

\definecolor{iccvblue}{rgb}{0.21,0.49,0.74}
\usepackage[pagebackref,breaklinks,colorlinks,allcolors=iccvblue]{hyperref}
\renewcommand{\red}[1]{\textbf{\textcolor{red}{#1}}}

\title{Improving Consistency in Diffusion Models for Image Super-Resolution}

\author{
    Junhao Gu$^{1,2}$\thanks{Work was done during interning at vivo. The first two authors share equal contributions. }
    ~~Peng-Tao Jiang$^{2*}$\thanks{Corresponding authors.}
    ~~Hao Zhang$^{2}$
    ~~Mi Zhou$^{2}$
    ~~Jinwei Chen$^{2}$ \\
    ~~Wenming Yang$^{1}$\footnotemark[2]
    ~~Bo Li$^{2}$
    \\
    $^1$Tsinghua University~~~ $^2$vivo Mobile Communication Co., Ltd \\
}

\begin{document}
\maketitle

\begin{abstract}
\label{Abstract}
% Real-world image super-resolution (Real-ISR) aims at restoring high-quality (HQ) images from low-quality (LQ) inputs corrupted by unknown and complex degradations.
% In particular, pretrained text-to-image (T2I) diffusion models provide strong generative priors to reconstruct credible and intricate details.
Recent methods exploit the powerful text-to-image (T2I) diffusion models for real-world image super-resolution (Real-ISR) and achieve impressive results compared to previous models.
However, we observe two kinds of inconsistencies in diffusion-based methods which hinder existing models from fully exploiting diffusion priors.
The first is the semantic inconsistency arising from diffusion guidance.
T2I generation focuses on semantic-level consistency with text prompts, while Real-ISR emphasizes pixel-level reconstruction from low-quality (LQ) images, necessitating more detailed semantic guidance from LQ inputs.
The second is the training-inference inconsistency stemming from the DDPM, which improperly assumes high-quality (HQ) latent corrupted by Gaussian noise as denoising inputs for each timestep.
To address these issues, we introduce ConsisSR to handle both semantic and training-inference consistencies.
On the one hand, to address the semantic inconsistency, we proposed a Hybrid Prompt Adapter (HPA). 
Instead of text prompts with coarse-grained classification information, we leverage the more powerful CLIP image embeddings to explore additional color and texture guidance.
On the other hand, we introduce Time-Aware Latent Augmentation (TALA) to bridge the training-inference inconsistency.
Based on the probability function p(t), we accordingly enhance the SDSR training strategy. With LQ latent with Gaussian noise as inputs, our TALA not only focuses on diffusion noise but also refine the LQ latent towards the HQ counterpart.
Our method demonstrates state-of-the-art performance among existing diffusion models.
The code will be made publicly available.
\end{abstract}

\section{Introduction}
\label{Introduction}

\begin{figure*}[t]
    \centering
    \setlength{\abovecaptionskip}{0pt}
    \begin{subfigure}{0.32\linewidth}
        \centering
        \includegraphics[width=\textwidth]{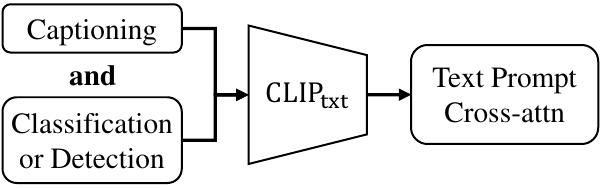}
        \caption{PASD}
        \label{fig:sub2}
    \end{subfigure}
    \hfill
    \begin{subfigure}{0.32\linewidth}
        \centering
        \includegraphics[width=\textwidth]{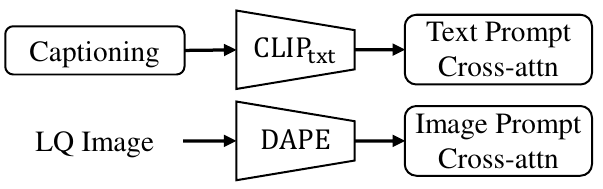}
        \caption{SeeSR}
        \label{fig:sub3}
    \end{subfigure}
    \hfill
    \begin{subfigure}{0.32\linewidth}
        \centering
        \includegraphics[width=\textwidth]{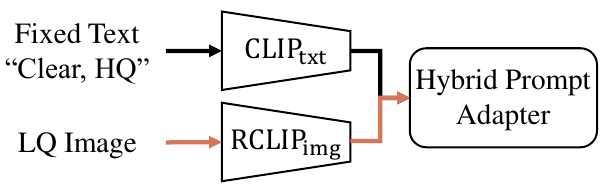}
        \caption{Ours}
        \label{fig:sub4}
    \end{subfigure}
    \caption{Comparisons among existing semantic prompts for SDSR.
    Previous SDSR methods only apply the SD prior to text prompts, such as captioning, whereas handling image prompts requires retraining additional layers.
    Instead, we leverage the consistent CLIP embedding space for both text and image prompts, which efficiently provide more fine-grained semantic guidance.
    }
    \label{fig:Semantic}
    \vspace{-10pt}
\end{figure*}

With the increasing prevalence of image-capturing devices in our daily lives, there emerges a growing need to capture clean, high-resolution images. Nevertheless, real-world images invariably contend with various degradation.
To address this issue, real-world image super-resolution (Real-ISR) techniques adeptly reconstruct the high-quality (HQ) image from the low-quality (LQ) input.

Various methods~\citep{dong2014learning, dong2015image, kim2016accurate, lim2017enhanced, zhang2018image, dai2019second, niu2020single} harness convolutional neural networks (CNN) to achieve remarkable performance, followed by transformer models~\citep{ liang2021swinir, zhang2022efficient}. Some others employ GANs' adversarial training to generate more photo-realistic images~\citep{ledig2017photo, wang2018esrgan}.
% However, most of them depend on basic bicubic down-sampling, limiting their efficacy in handling complex and unknown degradations encountered in real-world image super-resolution (Real-ISR).
However, most of the above methods assume LQ inputs with basic bicubic down-sampling, limiting their efficacy in handling complex and unknown degradations encountered in Real-ISR.
To tackle this problem, some methods manage to model the real-world degradations with complex degradation models, including degradation shuffle from BSRGAN~\citep{zhang2021designing} and high-order degradation from Real-ESRGAN~\citep{wang2021real}. 
Other GAN-based~\citep{liang2022details, liang2022efficient, chen2022real} or GAN-prior~\citep{menon2020pulse, pan2021exploiting, chan2021glean} models also achieve impressive performance.
While these methods demonstrates better perceptual quality, their generative capacity is still limited and the adversarial training in GANs is unstable, which may leads to unrealistic artifacts.

% Recently, emerging diffusion models~\citep{sohl2015deep, ho2020denoising, song2020score, dhariwal2021diffusion}, particularly pretrained large-scale text-to-image (T2I) models such as StableDiffusion (SD)~\citep{rombach2022high}, have exhibited superior performance in image generation against GANs.
%
Recently, emerging diffusion models~\citep{sohl2015deep, ho2020denoising, song2020score, dhariwal2021diffusion}, particularly large-scale text-to-image (T2I) models, such as StableDiffusion (SD)~\citep{rombach2022high}, have exhibited superior performance in image generation against GANs.
Later with ControlNet~\citep{zhang2023adding}, various SD-based super-resolution (SDSR) methods can leverage pre-trained diffusion priors and generate clear images under the fidelity constraint from LQ images.
%
% To better explore the semantic guidance, some SDSR methods~\citep{sun2024coser, yang2024pasd} 
% attempt to introduce the semantic information obtained from the LQ images 
% into the pretrained diffusion models.
%
% As shown in~\cref{fig:Semantic}(a), some of them~\citep{sun2024coser, yang2024pasd} involve image caption results, which focuses on coarse-grained classification information while neglecting the \blue{color or texture details from the LQ images.}
%
% Given the difference that T2I models tend to focus on semantic consistency 
% between the generated images and text prompts.
As shown in~\cref{fig:sub2}, some of them~\citep{sun2024coser, yang2024pasd} incorporate image captioning into UNet's cross attention layers, which leads T2I models to focus on semantic alignment between the SR results and text prompts.
However, SDSR methods focus on pixel-level reconstruction which necessitates more fine-grained semantic guidance from the LQ images.
To alleviate inconsistent semantic information requirements between T2I generation and SDSR tasks, SeeSR~\citep{wu2024seesr} additionally extracts soft image prompts from LQ images. But the misalignment between CLIP~\citep{radford2021learning} and their DAPE results in introducing additional attention layers.
Leveraging precise semantic guidance is inherently a complex task that necessitates extensive training.
Given CLIP's proficiency in joint text and image embedding, we introduce the Hybrid Prompt Adapter (HPA) to adapt pretrained T2I cross-attention layer for joint image and text prompts to ensure semantic consistency.
With fine-grained semantic guidance, our ConsisSR can circumvent undesirable artifacts and achieve superior perceptual quality.

\begin{figure}[h]
    \vspace{-10pt}
    \centering
    \setlength{\abovecaptionskip}{0pt}
    \includegraphics[width=\linewidth]{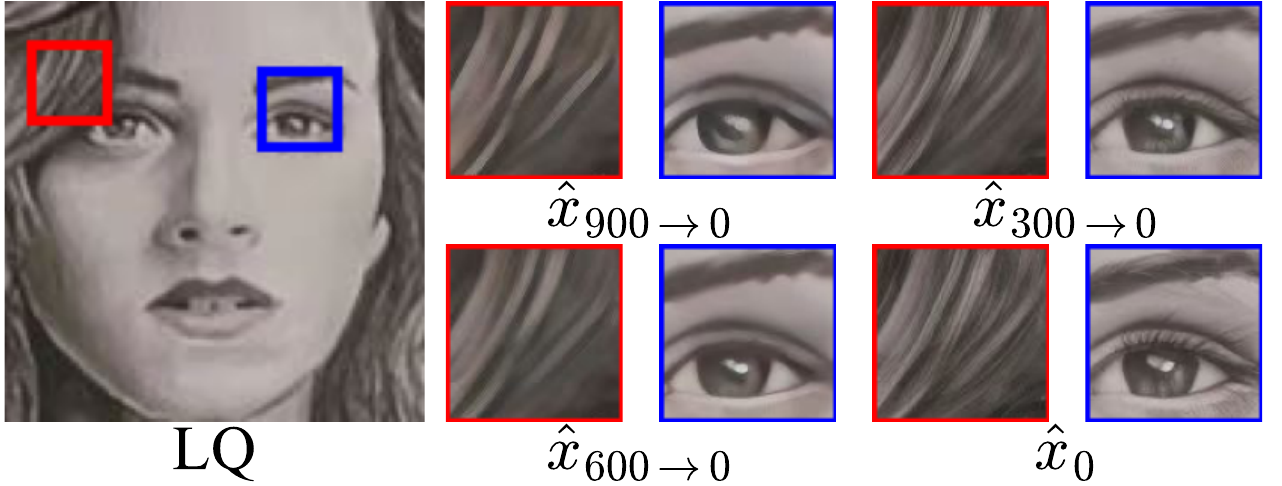}
    % \vspace{-10pt}
    \caption{Visualization of the truncated outputs reveals that smooth results naturally emerge in the early timesteps.
    This indicates that the denoising inputs gradually transition from LQ latent with Gaussian noise to that of HQ latent.}
    \label{fig:visual}
    \vspace{-10pt}
\end{figure}

Furthermore, we revisit the training process for these SDSR methods. DDPM~\citep{ho2020denoising} training assumes all inputs consist of HQ latent data corrupted by Gaussian noise, which only promotes the prediction of timestep-specific Gaussian noise and assumes the remaining input latent to be credible for each timestep.
We truncate the predicted $\hat{\mathbf{x}}_{t \rightarrow 0}$ at various timesteps and decode them into images as shown in~\cref{fig:visual}.
It is evident that in the early sampling timesteps (t $\rightarrow$ 1000), $\hat{\mathbf{x}}_{t \rightarrow 0}$ still appears overly smooth and noisy, which shows a clear discrepancy from the HQ distribution.
As DDPM training assumes $\hat{\mathbf{x}}_{t \rightarrow 0}$ to be accurate HQ latent, the discrepancies within the denoised latent would accumulate and ultimately compromise the reconstruction fidelity during the inference process.
Therefore, we propose Time-Aware Latent Augmentation (TALA) to ensure training-inference consistency.
By accordingly replacing HQ latent with LQ inputs in diffusion training, our denoising model not only eliminate timestep-specific Gaussian noise but also the residual between LQ latent and HQ latent.
In this way, we adapt our DM for LQ latent input and bridge the inconsistency between training assumption and inference inputs, improving sampling fidelity.

In this paper, we introduce ConsisSR to handle both semantic and training-inference consistency for T2I diffusion prior.
Specifically, we firstly propose HPA to incorporate the more powerful CLIP image embedding with text embedding, providing fine-grained semantic guidance. This prevents undesirable artifacts and improves semantic consistency.
Then we review the questionable assumption in DDPM training and introduce the TALA strategy. We intentionally corrupt the DM inputs in the early timesteps to enhance training-inference consistency, leading to improved fidelity performance.
% Our ConsisSR achieves SOTA results within SDSR methods.
Our contributions are as follows:
\begin{itemize}
% \itemindent=-13pt 
\item Compared to coarse-grained text descriptions, we integrate the more powerful CLIP image embedding, which encapsulates additional color and texture details.
Our Hybrid Prompt Adapter (HPA) simultaneously handles text and image prompts, adeptly leveraging T2I priors to enhance semantic consistency.

\item We review the conventional SDSR training that assumes HQ latent inputs.
In response, our Time-Aware Latent Augmentation (TALA) accordingly substitutes these HQ inputs with LQ ones in the early timesteps. This promotes the diffusion model to simultaneously address diffusion noise and the residuals between LQ and HQ latent, thereby improving training-inference consistency.

\item Extensive qualitative and quantitative experiments demonstrate the superiority of our ConsisSR with respect to the visual quality over the state-of-the-art methods.
\end{itemize}

\section{Related Work}
\label{Related}

\begin{figure*}[t]
    \setlength{\abovecaptionskip}{0pt}
    \centering
    \includegraphics[width=\linewidth]{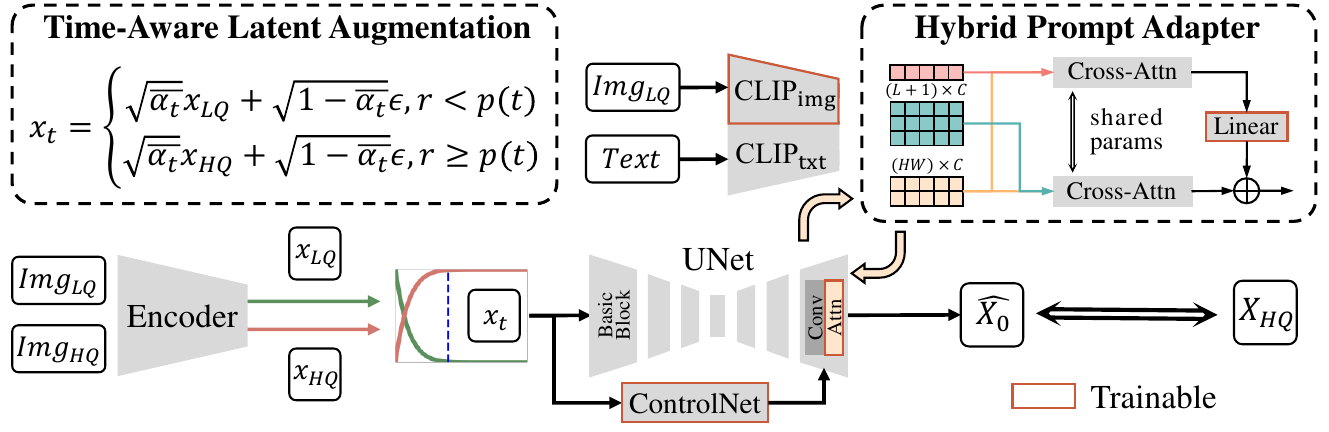}
    \caption{Overall training pipeline of our ConsisSR. 
    We propose the TALA strategy, which accordingly substitutes HQ inputs with LQ ones in the early timesteps, thereby improving training-inference consistency. Additionally, we introduce the HPA module to leverage both CLIP's text and image embeddings to enhance semantic consistency, thereby producing more credible textures.}
    \label{fig:overall}
    \vspace{-10pt}
\end{figure*}

\subsection{Real-World Image Super-Resolution}
While deep learning has made notable advancements in image super-resolution~\citep{dong2015image, lim2017enhanced, zhang2018image, dai2019second, niu2020single, liang2021swinir, zhang2022efficient}, most of them rely on simple bicubic degradation, which restricts their effectiveness in handling complex and unknown degradations in real-world scenarios.
% They also encounter challenges such as overly smoothed details when minimizing fidelity objectives. 
Real-world image super-resolution seeks to reconstruct photo-realistic image details.
Some works explore complex degradation models to approximate the real-world degradations~\citep{zhang2021designing, wang2021real}. 
Further in~\citep{menon2020pulse, pan2021exploiting, chan2021glean}, they leverage pretrained StyleGAN~\citep{karras2019style} as generative priors for Real-ISR.
Even though they excel at robustly removing degradation in Real-ISR tasks, their limited generative capacity often hinders them from generating realistic details.

Emerging diffusion models~\cite{dhariwal2021diffusion, rombach2022high} exhibit a remarkable capability to generate high-quality images.
For ISR tasks, several approach train their DMs from scratch on pixel space~\citep{choi2021ilvr, saharia2022image}.
While the former necessitates hundreds of diffusion steps, others~\citep{xia2023diffir, chen2024hierarchical} apply DM on compact latent space, but their generative ability is greatly restricted by the transformer backbone.
Resshift~\citep{yue2024resshift} constructs a Markov chain to shift the residual between LQ and HQ images with only 15 steps. Additionally, SinSR~\citep{wang2024sinsr} advances this strategy to single-step with consistency preserving distillation.
However, without strong diffusion priors, these methods still struggle to generate realistic and intricate textures.

\subsection{SD-based Super-Resolution}
For pretrained DMs, ControlNet~\citep{zhang2023adding} introduces an effective conditioning method, enabling broader applications.
Based on this, various SD-based super-resolution (SDSR) methods have achieved unprecedented success~\citep{wang2024exploiting, lin2023diffbir, sun2024coser, wu2024seesr, yang2024pasd, sun2023improving, wu2024one}.
Some of them~\citep{wang2024exploiting, lin2023diffbir, sun2023improving} do not apply semantic embedding to guide the diffusion process, leading to sub-optimal performance.
PASD~\citep{yang2024pasd} employ pretrained models including BLIP2~\citep{li2023blip} together with CLIP~\citep{radford2021learning} text encoder to provide semantic guidance.
Similarly, CoSeR~\citep{sun2024coser} train a cognitive encoder to approximate the cascaded outputs of BLIP2 and CLIP.
But they both involve the image captioning process, which focus on coarse-grained classification information while neglecting the color or texture details, as shown in~\cref{fig:CLIP}.
SeeSR~\citep{wu2024seesr} fine-tunes RAM~\citep{zhang2024recognize} model and separately integrates text prompt and image prompts with additional cross-attention layers.
However, given CLIP's proficiency in mapping text and image to a joint embedding space, the T2I priors from the SD model can also be applied to image prompts.
Our Hybrid Prompt Adapter harnesses our robust CLIP image encoder (RCLIP) and exploits the domain alignment in CLIP embeddings.
This approach adapts SD's cross-attention layers for both text and image prompts, thereby enhancing semantic consistency.

Furthermore, the aforementioned SDSR methods follow the training protocol from DDPM~\citep{ho2020denoising}.
Its drawback lies in assuming all inputs as HQ latent data with Gaussian noise.
Therefore, we accordingly corrupt the DM inputs in the early timesteps to improve training-inference consistency, resulting in better fidelity performance.
% enable our ConsisSR to refine the accumulated latent representations and improve training-inference consistency.

\section{Method}
\label{Method}

\subsection{Preliminary: diffusion models}

Diffusion models are probabilistic models designed to generate data samples by gradually denoising a normally distributed variable $\mathbf{x}_T \sim  \mathcal{N}(0, 1)$ in $T$ iteration steps.
In each forward iteration, a Gaussian noise with variance $1-\alpha_t$ is added to $\mathbf{x}_{t-1}$, and the overall forward process can be described as:
\begin{equation}
\mathbf{x}_t = \sqrt{\bar \alpha_t} \mathbf{x}_0 + \sqrt{1-\bar \alpha_t} \epsilon, 
\end{equation}
where $\bar \alpha_t = \textstyle\prod_{i=0}^{t} \alpha_i$ and $\epsilon \sim  \mathcal{N}(0, 1)$.
For the reverse process, diffusion models first sample a Gaussian noise $\mathbf{x}_T$ as the start point. Subsequently, conditioning on $c$, they iteratively estimate the added noise for each step $t$ through the denoising network $\epsilon_\phi$ until reaching the clean output $\hat{\mathbf{x}}_0$. 
The optimization objective is defined as follows:
\begin{equation}
\mathcal{L}_{DM} = \mathbb{E}_{\mathbf{x}, \epsilon \sim  \mathcal{N}(0, 1), t}\left [ \left \| \epsilon - \epsilon_\phi (\mathbf{x}_t, t, c) \right \|_2^2 \right ].
\end{equation}

\subsection{Overall Pipeline}
As shown in~\cref{fig:overall}, we demonstrate our training augmentation on the left and the network architecture on the right.
Firstly, we equip the denoising UNet with ControlNet~\citep{zhang2023adding} to manipulate the output using LQ images.
Subsequently, we put forward the Time-Aware Latent Augmentation (TALA) that accordingly substitutes the HQ inputs with LQ ones in the early timesteps. This enhance training-inference consistency, leading to superior fidelity performance.
Furthermore, we introduce the Hybrid Prompt Adapter (HPA), which leverages both CLIP's text and image embeddings as semantic guidance in a decoupled manner, thereby generating more realistic and credible texture.
These approaches enable our model to bridge the inherent gaps between T2I generation and Real-ISR tasks.

\subsection{Hybrid Prompt Adapter}
SD originates from the T2I generation task, hence original cross-attention is only tailored to text prompts. However, given that CLIP can inherently map images and text to a joint latent space, we intuitionally suggest that we can leverage the LQ image as fine-grained semantic guidance.
To align the SD backbone with Real-ISR, we propose the Hybrid Prompt Adapter (HPA) to incorporate both input image prompt and text prompt to enhance semantic consistency.

\begin{figure}[ht]
    \centering
    % \vspace{-10pt}
    % \setlength{\abovecaptionskip}{5pt}
    \includegraphics[width=\linewidth]{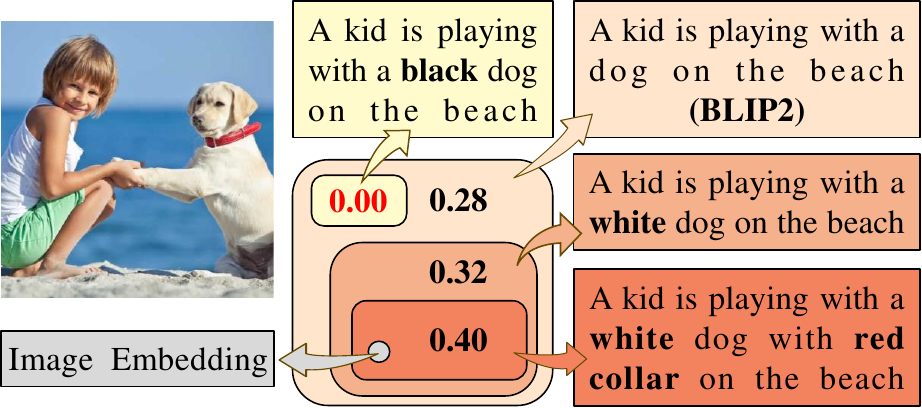}
    % \vspace{-10pt}
    \caption{Venn diagram of CLIP's embedding space. The similarity is derived from the cosine distance after Softmax between image and various text descriptions. }
    \label{fig:CLIP}
    \vspace{-10pt}
\end{figure}

\begin{figure}[t]
    \centering
    \includegraphics[width=\linewidth]{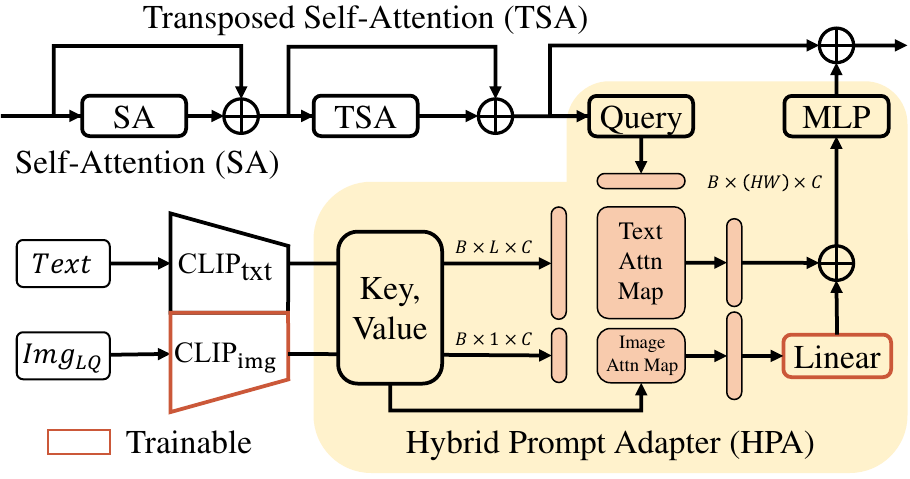}
    % \vspace{-10pt}
    \caption{Network architecture of our transformer block with the proposed Hybrid Prompt Adapter (HPA).}
    \label{fig:HPA}
    % \vspace{-10pt}
\end{figure}

Although other SDSR methods also attempt to incorporate semantic guidance, including CoSeR~\citep{sun2024coser}, PASD~\citep{yang2024pasd}, they both involve the image captioning process, which focus on coarse-grained classification information while neglecting the color or texture details.
As shown in~\cref{fig:CLIP}, we present various text descriptions in a Venn diagram pattern. Following~\citet{radford2021learning}, we calculate the cosine distance between the CLIP image embedding and each text embedding, and gauge the similarity between the input image and each text description based on the results after the Softmax function.
It can be observed that while BLIP2 offers a reasonable caption, it falls short in capturing the detailed information like color and accessory (e.g., white dog with red collar in~\cref{fig:CLIP}).
The most precise and elaborate description exhibits the highest similarity, indicating the intricate details inherent in the image embedding itself.
Besides, T2I-Adapter~\cite{mou2023t2i} and IP-Adapter~\cite{ye2023ip} also have shown that image prompts can provide more accurate control over color or structure than text prompts in image generation task.
Therefore, we incorporate the more powerful CLIP image embedding with text prompt in our HPA, providing fine-grained semantic guidance for our denoising network, as depicted in~\cref{fig:HPA}.

Firstly, for the CLIP image encoder, we conduct a straightforward fine-tuning process to enhance its robustness to LQ images while preserving its alignment with the CLIP embedding domain.
\begin{align}
\mathcal{L}_{CLIP} = \mathbb{E} \left [ \left \| \widehat{\operatorname{CLIP_{img}}}(\mathbf{x}_{HQ}) - \operatorname{RCLIP_{img}}(\mathbf{x}_{LQ}) \right \|_2^2 \right ],
\end{align}
where $\widehat{\operatorname{CLIP_{img}}}$ represents fixed original CLIP image encoder and $\operatorname{RCLIP_{img}}$ represents our fine-tuned robust CLIP image encoder.

Then, the text prompt and low-quality (LQ) image are separately fed into the original CLIP text encoder and our robust CLIP image encoder, resulting in text and image prompt embeddings. 
Both prompt embeddings are subsequently transformed into Key and Value vectors using the original linear layers, which are then used to compute attention maps with Query vectors derived from the image latent.
Akin to~\citet{ye2023ip}, we employ a decoupled cross-attention, which involves independently processing the interactions between two pairs of Key and Value vectors with the Query vector.
Differently, to preserve the SD prior as much as possible, we keep all projection layers fixed in our HPA for both denoising UNet and ControlNet.
Correspondingly, the output of the additional image prompt branch is added to the original text prompt branch after 
passing through a trainable zero-initialized linear layer.

Apart from HPA, we also made minor adjustments to the self-attention layers.
For the original self-attention, we incorporate learned absolute positional embeddings to better align with the image domain.
And we introduce transposed self-attention inspired by the work in~\cite{zamir2022restormer}, which helps to further model global context.
For the self-attention layers, we only train the added parameters in the denoising UNet. 

\begin{figure*}[ht]
    \centering
    \setlength{\abovecaptionskip}{0pt}
    \includegraphics[width=\linewidth]{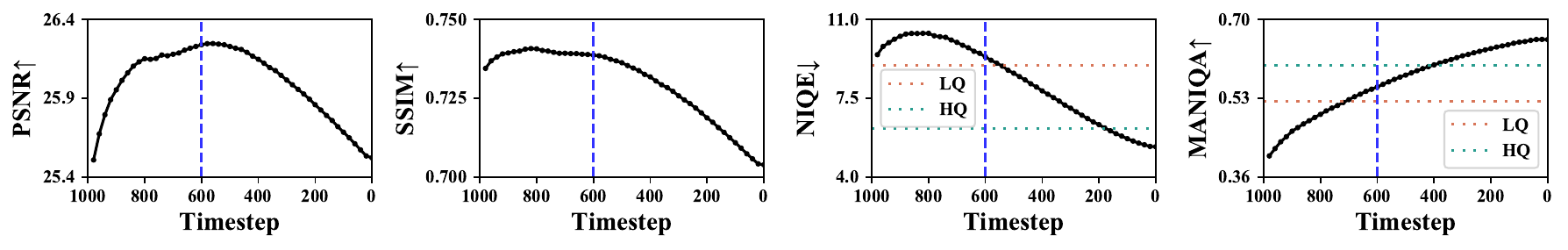}
    % \vspace{-10pt}
    \caption{IQA metrics of the truncated outputs at different diffusion steps. Fidelity metrics represented by PSNR and SSIM on the left, while perceptual quality metrics represented by NIQE and MANIQA on the right.}
    \label{fig:metric}
    \vspace{-10pt}
\end{figure*}

\subsection{Time-Aware Latent Augmentation}
\label{sec:TALA}

Although diffusion models are able to infer detailed and credible images, the truncated outputs in the early timesteps exhibit excessive smoothness and noise, as shown in \cref{fig:visual}.
However, DDPM training assumes HQ latent with Gaussian noise as inputs for each timestep.
This discrepancies between training assumption and inference inputs would inherently compromises sampling fidelity.
To tackle this problem, we propose the Time-Aware Latent Augmentation (TALA) to improve training-inference consistency.

% For T2I generation tasks, there are no exact requirements for detailed composition, such as the relative size and position of different objects. However, for Real-ISR tasks, it is essential to prioritize structural consistency over blindly pursuing perceptual quality.
% Although ControlNet~\citep{zhang2023adding} introduces an effective conditioning method to provide pixel-level guidance, most SDSR methods still overlook drawbacks within DDPM training.

To take a deeper insight, we take our ConsisSR as an example and truncate the denoised $\hat{x}_{t \rightarrow 0}$ at each step and decode them into images. We further quantified their reconstruction quality over timesteps in the above~\cref{fig:metric}.
We can clearly observe that sampling fidelity, represented by PSNR and SSIM, remarkably improves and remains stable until 40\% (timestep 600). 
Subsequently, at the cost of fidelity, the diffusion model steadily enhances texture details, leading to an improvement in the perceptual quality indicated by NIQE and MANIQA.
Moreover, we can observe that the truncated results do not exhibit superior perceptual quality over the LQ inputs until timestep 600.
This implies that in the early timesteps, the inputs to the denoising network lean towards LQ latent corrupted by Gaussian noise instead of DDPM's HQ latent assumption.

\vspace{-5pt}
\begin{algorithm}[ht]
    
    \caption{Time-Aware Latent Augmentation}
    \label{alg:train}
    \begin{algorithmic}[1]
    \REQUIRE Training set ($\mathbf{X}_{LQ}$, $\mathbf{X}_{HQ}$), Text Prompt $Text$
    \WHILE{not converged}
        \STATE sample $\mathbf{x}_{LQ}$, $\mathbf{x}_{HQ}$ from ($\mathbf{X}_{LQ}$, $\mathbf{X}_{HQ}$)
        \STATE sample $t \sim \mathcal{U}(\{1, ..., T\})$
        \STATE sample $\epsilon \sim \mathcal{N}(0,1)$
        \STATE $c_{img} = RCLIP_{img}(x_{LQ})$
        \STATE $c_{txt} = CLIP_{txt}(Text)$
        \STATE sample $r \sim \mathcal{U}(0,1)$
        \STATE $\mathbf{x}_t =\left\{\begin{matrix}
                \sqrt{\bar{\alpha}_t} \mathbf{x}_{LQ} + \sqrt{1-\bar{\alpha}_t} \epsilon, r < p(t)\\ 
                \sqrt{\bar{\alpha}_t} \mathbf{x}_{HQ} + \sqrt{1-\bar{\alpha}_t} \epsilon, r \geq  p(t)\\ 
                \end{matrix}\right.$
        \STATE $\hat{\epsilon} = \epsilon_\phi(\mathbf{x}_t, t, c_{img}, c_{txt})$
        % \STATE $\tilde{\epsilon} = \frac {{\sqrt{\bar{\alpha_t}}}x_{HQ} - x_t}{\sqrt{1-\bar{\alpha_t}}}$
        \STATE Update model with $\left \| \hat{\epsilon} - \frac {\mathbf{x}_t - {\sqrt{\bar{\alpha}_t}}\mathbf{x}_{HQ}}{\sqrt{1-\bar{\alpha}_t}} \right \|_2^2$
    \ENDWHILE
    % \UNTIL converged
    \end{algorithmic}
\end{algorithm}
\vspace{-5pt}

Therefore, to enhance training-inference consistency, we propose TALA strategy which intentionally corrupt the DM inputs during the early timesteps.
As shown in~\cref{alg:train}, instead of the DDPM approach, which follows an HQ+Gaussian to HQ denoising pipeline, we enhance the diffusion training by employing an LQ+Gaussian to HQ denoising pipeline.
We devise a time-dependent probability function $p(t)$ to handle augmentation strength.
For each timestep $t$, we probabilistically substitude the input from HQ latent corrupted by Gaussian noise with the LQ one according to $p(t)$, as shown in Line 8 in~\cref{alg:train}.
Based on the afore-mentioned IQA results over timesteps, we can divide the inference process into two phases: the former phase, before 600 timesteps, focuses on improving fidelity, while the latter phase, after 600 timesteps, is oriented towards enhancing perceptual quality.
Hence, we regard timestep 600 as the ending point, and our TALA strategy primarily operates before this timestep. We set the starting point $p(1000)$ to be fixed at 1 and control $p(600)$ to be below $1\%$. Empirically, we adopt a power function $p(t) = (t / T)^{\gamma}$ and obtain $\gamma=10$. 
For ablation studies on probability function $p(t)$, please refer to~\cref{para:TALA}.

\begin{figure*}[t]
    \setlength{\abovecaptionskip}{0pt}
    \centering
    \includegraphics[width=\linewidth]{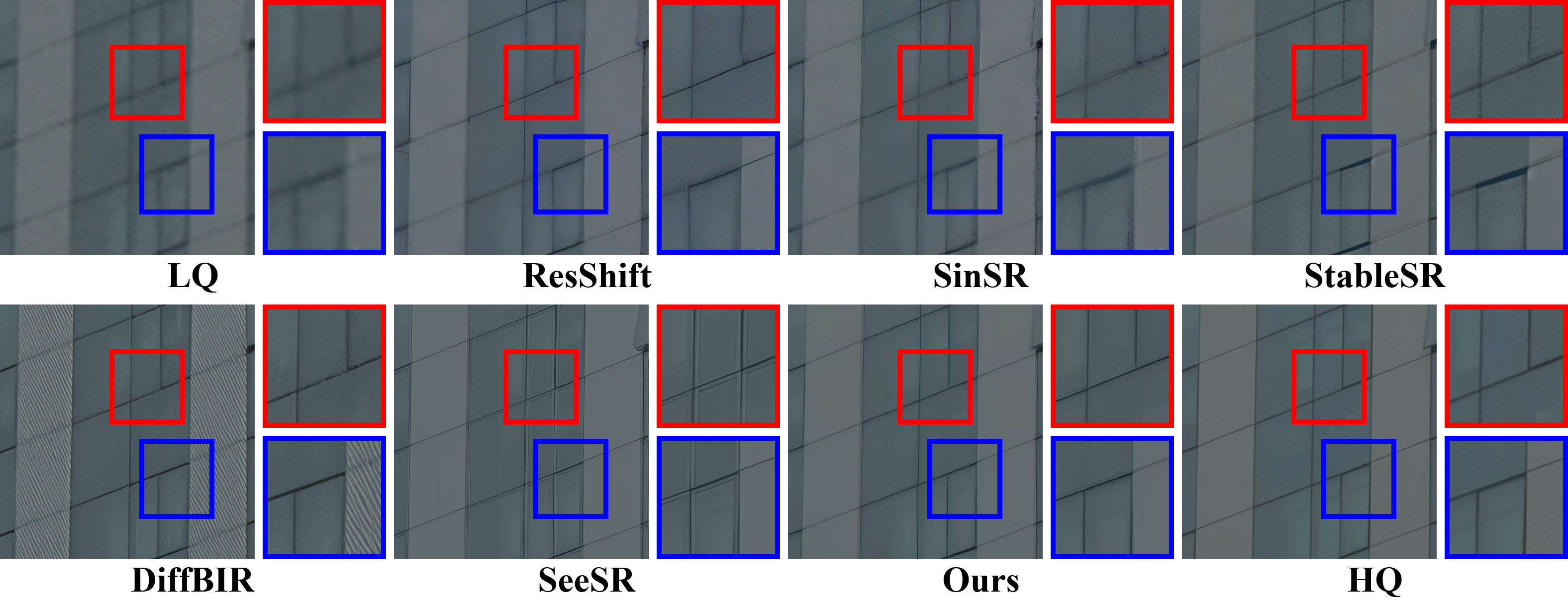}
    % \vspace{5pt}
    \includegraphics[width=\linewidth]{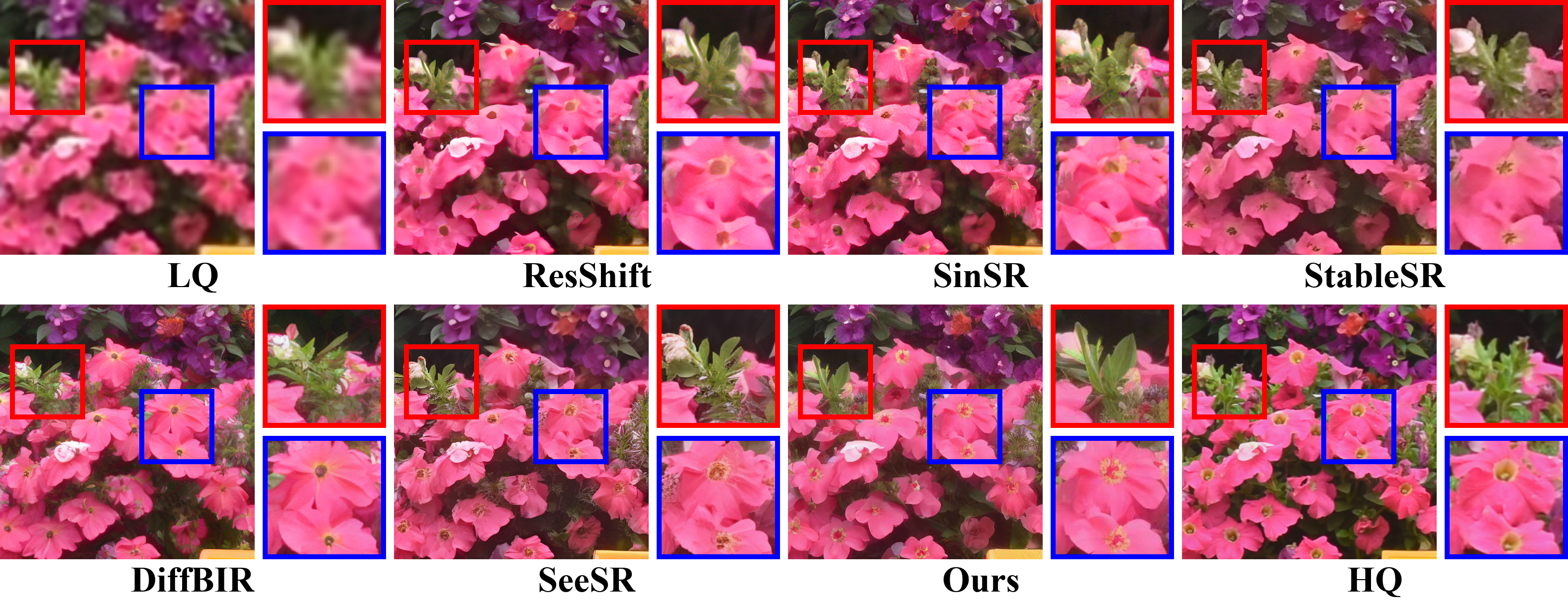}
    \caption{Visual comparisons among different Real-ISR methods. 
    Please zoom in for a better view.} 
    \label{fig:SOTA}
    \vspace{-15pt}
\end{figure*}

Following this strategy, our supervision is no longer solely timestep-specific diffusion noise but also includes residuals between LQ and HQ latents.
When $r < p(t)$, our TALA is enabled and our training objective in Line.~10 can be further formulated as: 
\begin{align}
\left \| \hat{\epsilon} - \frac {\mathbf{x}_t - {\sqrt{\bar{\alpha}_t}}\mathbf{x}_{HQ}}{\sqrt{1-\bar{\alpha}_t}} \right \|_2^2 = \left \| \hat{\epsilon} - (\epsilon + \frac {\sqrt{\bar{\alpha}_t} (\mathbf{x}_{LQ} - \mathbf{x}_{HQ})}{\sqrt{1-\bar{\alpha}_t}} )\right \|_2^2.
\end{align}
This promotes the denosing network to not only remove diffusion noise  but also refine the LQ latent towards the HQ counterpart.
Consequently, our TALA enables SDSR models to effectively handle LQ Latent inputs in the early timesteps and improve training-inference consistency, with minimal interference to the original DDPM training.

\section{Experiments}
\label{Experiment}

\subsection{Experimental Settings}
\label{subsec:expset}

% \vspace{-10pt}
\paragraph{Datasets.}
% We train ConsisSR on our filtered ImageNet~\citep{deng2009imagenet} with around 18M images. 
We train ConsisSR on the ImageNet~\citep{deng2009imagenet} dataset.
We employ the degradation pipeline from Real-ESRGAN~\citep{wang2021real} to generate LQ-HQ training pairs.
For testing datasets, we adopt the widely used DIV2K-Val~\citep{agustsson2017ntire} as the synthetic dataset, along with RealSR~\cite{cai2019toward} and DrealSR~\cite{wei2020component} as the real-world datasets.

\vspace{-10pt}
\paragraph{Evaluation Metrics.}
For quantitative evaluation of Real-ISR models, we first adopt full-reference metrics, including PSNR (calculated on the RGB channels), SSIM (calculated on the Y channel), and LPIPS~\citep{zhang2018unreasonable} for fidelity evaluation with HQ targets.
Then for perceptual quality, we adopt the no-reference metrics, including NIQE~\citep{mittal2012making}, MANIQA~\citep{yang2022maniqa}, MUSIQ~\citep{ke2021musiq} and CLIPIQA~\citep{wang2023exploring}.

\vspace{-10pt}
\paragraph{Implementation details.}
For HPA, we crop central patches from the ImageNet images and resize them to $224 \times 224$ for robust CLIP fine-tuning.
Both horizontal and vertical flips are performed for data augmentation.
The CLIP image encoder is trained with 2 NVIDIA L40s GPUs and the batch size is set to 8 per GPU.
We adopt Adam as optimizer ($\beta_1$ = 0.9, $\beta_2$ = 0.99), and we train the model for 100K iterations with a learning rate fixed at $1 \times 10^{-5}$.

For SDSR training, the Stable Diffusion 2.1-base is used as the pretrained T2I model, which is the same as~\citep{wang2024exploiting, lin2023diffbir}.
We crop central patches of size $512 \times 512$ from ImageNet.
Both horizontal and vertical flips are performed for data augmentation.
The batch size is set to 2 per GPU, totaling 8 with 4 NVIDIA L40s.
We train it for 200K iterations with the Adam optimizer ($\beta_1$ = 0.9, $\beta_2$ = 0.99) and the learning rate is fixed at $5 \times 10^{-5}$.
And we adopt the spaced DDPM sampling~\citep{nichol2021improved} for 50-step inference.

\begin{table*}[t]
    \centering
    \fontsize{12pt}{12pt}\selectfont
    \resizebox{\linewidth}{!}{
        \begin{tabular}{c | c | c c c | c c c c | c}
            \toprule
            \multirow{2}{*}{Datasets} & \multirow{2}{*}{Methods} & \multicolumn{3}{c|}{Full-reference IQA} & \multicolumn{4}{c|}{No-reference IQA} & \multirow{2}{*}{Rank(avg)$\downarrow$}\\
              &  & PSNR$\uparrow$ & SSIM$\uparrow$ & LPIPS$\downarrow$ & NIQE$\downarrow$  & MANIQA$\uparrow$ & MUSIQ$\uparrow$ & CLIPIQA$\uparrow$ & \\
            \midrule
            \multirow{8}{*}{DrealSR} & ResShift &\blue{28.46} & \blue{0.7673} & 0.4006 & 8.1249 & 0.4586 & 50.60 & 0.5342 & 5.43\\
             & SinSR & 28.36 & 0.7515 & 0.3665 & 6.9907 & 0.4884 & 55.33 & 0.6383 & 5.00\\
             % & CCSR & \red{28.96} & \red{0.7710} & \red{0.2922} & \blue{5.9100} & 0.5682 & 58.82 & 0.5467 & \blue{3.14}\\
             & StableSR & 28.03 & 0.7536 & \blue{0.3284} & 6.5239 & 0.5601 & 58.51 & 0.6356  & 4.57\\
             & DiffBIR & 26.71 & 0.6571 & 0.4557 & \blue{6.3124} & 0.5930 & 61.07 & 0.6395 & 5.00\\
             & PASD & 27.36 & 0.7073 & 0.3760 & \red{5.5474} & \blue{0.6169} & 64.87 & \blue{0.6808} & 3.57\\
             & SeeSR & 28.17 & \red{0.7691} & \red{0.3189} & 6.3967 & 0.6042 & \blue{64.93} & 0.6804 & \blue{2.57}\\
             & Ours & \red{28.47} & 0.7581 & 0.3463 & 6.3668 & \red{0.6224} & \red{65.28} & \red{0.6965} & \red{1.86}\\
    
            \midrule
            \multirow{8}{*}{RealSR} & ResShift & \red{26.31} & \red{0.7421} & 0.3460 & 7.2635 & 0.5285 & 58.43 & 0.5444 & 5.14\\
             & SinSR & \blue{26.28} & \blue{0.7347} & 0.3188 & 6.2872 & 0.5385 & 60.80 & 0.6122 & 4.43\\
             % & CCSR & 26.24 & \blue{0.7365} & \red{0.2559} & 5.7400 & 0.5974 & 63.64 & 0.5287 & 4.43\\
             & StableSR & 24.70 & 0.7085 & \blue{0.3018} & 5.9122 & 0.6221 & 65.78 & 0.6178 & 4.57\\
             & DiffBIR & 24.75 & 0.6567 & 0.3636 & 5.5346 & 0.6246 & 64.98 & 0.6395 & 5.29\\
             & PASD & 25.21 & 0.6798 & 0.3380 & 5.4137 & \blue{0.6487} & 68.75 & \blue{0.6620} & 3.57\\
             & SeeSR & 25.18 & 0.7216 & \red{0.3009} & \blue{5.4081} & 0.6442 & \red{69.77} & 0.6612 & \blue{2.57}\\
             & Ours & 25.51 & 0.7033 & 0.3223 & \red{5.2655} & \red{0.6552} & \blue{69.48} & \red{0.6925} & \red{2.43}\\
    
            \midrule
            \multirow{8}{*}{DIV2K} & ResShift & \red{24.65} & \red{0.6181} & 0.3349 & 6.8212 & 0.5454 & 61.09 & 0.6071 & 4.86\\
             & SinSR & \blue{24.41} & 0.6018 & 0.3240 & 6.0159 & 0.5386 & 62.82 & 0.6471 & 4.86\\
             % & CCSR & \blue{24.46} & \blue{0.6113} & \red{0.3045} & \blue{4.6100} & 0.5912 & 62.78 & 0.5367 & 4.00\\
             & StableSR & 23.26 & 0.5726 & \red{0.3113} & 4.7581 & 0.6192 & 65.92 & 0.6771 & 4.00\\
             & DiffBIR & 23.64 & 0.5647 & 0.3524 & \blue{4.7042} & 0.6210 & 65.81 & 0.6704 & 4.71\\
             & PASD & 23.14 & 0.5505 & 0.3571 & \red{4.3617} & \red{0.6483} & \blue{68.95} & 0.6788 & 4.00\\
             & SeeSR & 23.68 & \blue{0.6043} & 0.3194 & 4.8102 & 0.6240 & 68.67 & \blue{0.6936} & \blue{3.00}\\
             & Ours & 23.95 & 0.5896 & \blue{0.3145} & 4.8323 & \blue{0.6433} & \red{69.13} & \red{0.7153} & \red{2.57}\\
             \bottomrule
        \end{tabular}
    }
    \caption{Comparison with SOTA methods. The best results are highlighted in red and the second best results are highlighted in blue.}
    \label{tab:SOTA}
    \vspace{-10pt}
\end{table*}

\subsection{Comparisons with SOTA methods}

We compare our ConsisSR with diffusion-based SR methods, which are split into two groups:
The first group focuses on conventional diffusion models without pretrained diffusion prior, which include ResShift~\citep{yue2024resshift} and SinSR~\citep{wang2024sinsr}.
The second group, comprising SDSR models, primarily aims to generate realistic and detailed textures, which includes, StableSR~\citep{wang2024exploiting}, DiffBIR~\citep{lin2023diffbir}, PASD~\citep{yang2024pasd} and SeeSR~\citep{wu2024seesr}.

The quantitative results for these methods are presented in~\cref{tab:SOTA}, and their average rank is listed in the last column.
While ResShift, SinSR achieve impressive fidelity as represented by full-reference metrics, their perceptual quality is apparently inferior to other methods.
DiffBIR and PASD achieve decent perceptual quality, but noticeably sacrifice fidelity.
In contrast, our method strikes a balance and excels in both aspects, achieving the best average rank across all three datasets.
This comparison highlights the effectiveness of our model in handling Real-ISR tasks.

To emphasize the superior performance of our approach, we provide qualitative comparisons in~\cref{fig:SOTA}.
For distinct and regular textures such as glass curtain walls, our approach can produce sharper and well-aligned edges while preventing excessive artifact generation in smooth regions. 
For intricate and random textures such as foliage and flowers, our method can also generate convincing and realistic texture details.
These comparisons vividly illustrate that our method achieves superior SR results, marked by clearer textures and sharper edges.
Please refer to our supplementary material for more quantitative comparison against diffusion acceleration methods and other diffusion prior model.

We further compare the model complexity of our ConsisSR with other SDSR methods.
We present the number of inference steps, computational cost and number of parameters in \cref{tab:Complexity}.
\begin{table}[htbp]
    \centering
    \scriptsize
    \setlength{\abovecaptionskip}{0pt}
    \resizebox{\linewidth}{!}{
        \begin{tabular}{c | c c c c c}
            \toprule
            Method & StableSR & DiffBIR & PASD & SeeSR & Ours\\
            %\midrule%
            \hline
            Steps & 200 & 50 & 20 & 50 & 50\\
            MACs(T) & 79.94 & 24.23 & 29.12 & 65.86 & 49.47\\
            Params(G) & 1.41 & 1.72 & 1.90 & 2.52 & 2.13\\
            \bottomrule
        \end{tabular}
    }
    \caption{Analysis of Model Complexity. Tested with an input image of size $512 \times 512$.}
    \label{tab:Complexity}
    \vspace{-10pt}
\end{table}
According to \cref{tab:SOTA} and \cref{tab:Complexity}, our model achieves lower computational cost and fewer parameters compared to the second-best performing model, SeeSR. Besides, when compared to other SDSR methods, although our model incurs higher computational overhead, it demonstrates notably superior performance.

\subsection{Ablation study}

We further implement several variants to demonstrate the effectiveness of each component in our model.
Unless otherwise specified, our ablation models adopt the same training settings as our ConsisSR model.
And we report PSNR and LPIPS as fidelity metrics, along with NIQE and CLIPIQA as perceptual quality metrics on the RealSR dataset.

% \vspace{-10pt}

\begin{figure*}[ht]
    \setlength{\abovecaptionskip}{0pt}
    \centering
    \includegraphics[width=\linewidth]{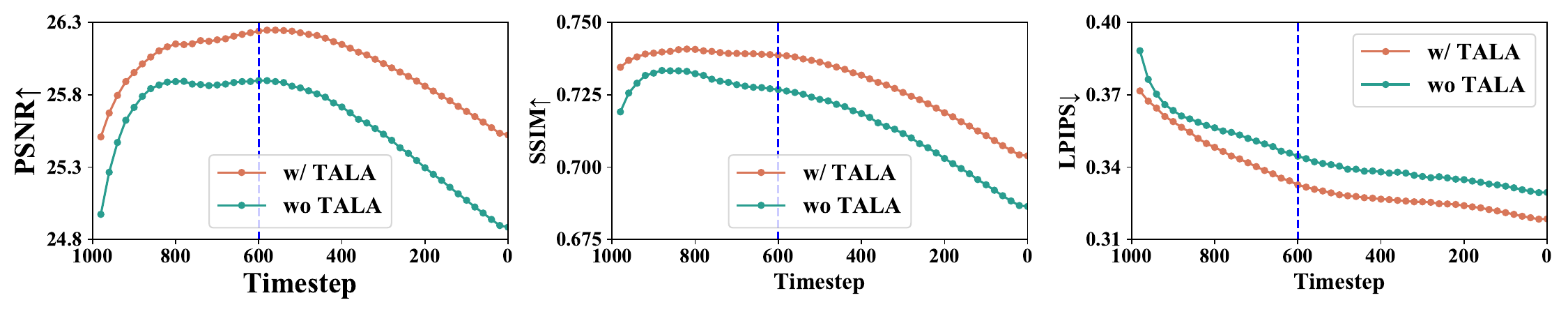}
    \caption{Full-reference IQA metrics over timesteps with and without TALA. Our TALA can significantly improve the sampling fidelity.}
    \label{fig:TALA}
    \vspace{-10pt}
\end{figure*}
% \vspace{-10pt}

\paragraph{Effectiveness of our prompt extractor HPA.}
% Firstly, we make minor adjustments to the self-attention layers by integrating learned absolute positional embeddings into the original layers and introducing transposed self-attention to better capture global context.
% As depicted in the upper part of~\cref{tab:HPA}, these modifications led to a 0.4dB increase in PSNR with a slight drop in generation quality.
Exploring precise semantic guidance is inherently a complex task that requires extensive training.
To ensure fair comparison, We adopt the same text prompt as the baseline for our experiments.
We compare with two other image prompt extractor: the IP-Adapter~\citep{ye2023ip} from image-to-image generation task and the DAPE soft prompt layers from SeeSR~\citep{wu2024seesr}.
And we also compare our HPA's decouple design with separating attention layers.

As pixel-level metrics like PSNR and SSIM favor blurry images, they are inadequate for assessing semantic consistency.
We adopt CLIP cosine distance $\operatorname{CLIP-I}$ from Dreambooth~\cite{ruiz2023dreambooth} as our evaluation metric.
To clearly demonstrate the relative improvement of semantic consistency during the SR process, we propose the CLIP improvement as follows:
\begin{align}
{\operatorname{\Delta CLIP-I}} = \frac {\operatorname{CLIP-I}(SR, HR)} {\operatorname{CLIP-I}(LR, HR)} - 1.
\end{align}
For fair comparison, we adopt Vit-B/32 as CLIP backbone, which differs from that of SD.
% \begin{table}[htbp]
%     \centering
%     % \setlength{\abovecaptionskip}{5pt}
%     \resizebox{\linewidth}{!}{
%         \begin{tabular}{c | c c c}
%             \toprule
%             Blocks & SA(baseline) & +PE & +PE+TSA\\
%             \midrule
%             PSNR$\uparrow$ & 24.76 & 24.91 & \red{25.16}\\
%             LPIPS$\downarrow$ & 0.3420 & 0.3482 & \red{0.3418}\\
%             NIQE$\downarrow$ & 5.5183 & \red{5.3348} & 5.5665\\
%             CLIPIQA$\uparrow$ & 0.6815 & \red{0.6889} & 0.6783\\
%             \midrule
%             Prompts & +IP-Adapter & +DAPE(soft) & +HPA\\
%             \midrule
%             PSNR$\uparrow$ & 24.41 & 25.07 & \red{25.50}\\
%             LPIPS$\downarrow$ & 0.3560 & 0.3219 & \red{0.3206}\\
%             NIQE$\downarrow$ & 5.4109 & 5.4126 & \red{5.3103}\\
%             CLIPIQA$\uparrow$ & \red{0.6947} & 0.6878 & 0.6917\\
%             $\Delta$CLIP-I$\uparrow$ & 0 & 0 & \red{0}\\
            
%             \bottomrule
%         \end{tabular}
%     }
%     \caption{Ablation studies on our transformer block. The results of the self-attention layers are displayed above, and the results of prompt extractors are shown below. The best results are highlighted in red.}
%     \label{tab:HPA}
%     % \vspace{-15pt}
% \end{table}

The results are listed in~\cref{tab:HPA}.
We can observe that IP-Adapter prioritizes perceptual quality over fidelity. This is primarily because its generation task pays attention to semantic similarity rather than pixel-level fidelity in Real-ISR.
Regarding DAPE, its overall performance also falls short of our HPA.
Additionally, separate cross-attention also proves effective for our HPA but introduces extra parameters and computational overhead.
For CLIP improvement, introducing image prompt proves beneficial for semantic consistency.
% The exploration of precise semantic guidance is inherently a complex task that requires extensive training.
Our HPA's superiority lies in leveraging the joint embedding space of CLIP to adapt the cross-attention prior for the image prompt, thereby enhancing the semantic consistency of the generated results.
This effectively improves semantic consistency and achieves superior perceptual quality as well as fidelity.
\begin{table}[htbp]
    \centering
    \setlength{\abovecaptionskip}{5pt}
    \resizebox{\linewidth}{!}{
        \begin{tabular}{c | c c | c c | c}
            \toprule
            \multirow{2}{*}{Methods} & \multicolumn{2}{c|}{Full-reference IQA} & \multicolumn{2}{c|}{No-reference IQA} & \multirow{2}{*}{$\Delta$CLIP-I$\uparrow$} \\
            & PSNR$\uparrow$ & LPIPS$\downarrow$ & NIQE$\downarrow$ & CLIPIQA$\uparrow$ &\\
            \midrule
            Fixed Text & 25.16 & 0.3418 & 5.5665 & 0.6783 & 0.47\%\\
            +IP-Adapter & 24.41 & 0.3560 & \blue{5.4109} & \red{0.6947} & 0.93\%\\
            +DAPE(soft) & 25.07 & \blue{0.3219} & 5.4126 & 0.6878 & 1.29\%\\
            % +DAPE & 25.11 & \blue{0.3210} & 5.4233 & 0.6807\\
            +HPA(separate) & \blue{25.37} & 0.3224 & 5.4233 & 0.6837 & \red{1.51\%}\\
            % \bottomrule
            +HPA(ours) & \red{25.50} & \red{0.3206} & \red{5.3103} & \blue{0.6917} & \blue{1.47\%}\\
            \bottomrule
        \end{tabular}
    }
    \caption{Ablation studies on our semantic prompt extractors. The best results are highlighted in red and the second best results are highlighted in blue.}
    \label{tab:HPA}
    \vspace{-10pt}
\end{table}

\paragraph{Effectiveness of our TALA training strategy.}
\label{para:TALA}
Our TALA is primarily introduced to enhance the training-inference consistency and improve sampling fidelity.
Therefore, we demonstrate the full-reference metrics of our model with respect to timesteps with and without TALA, as shown in~\cref{fig:TALA}.
We repeat ten times for each parameter setting and calculate the mean value. 
It can be observed that during the early timesteps, our model trained with TALA achieved remarkable advantages in terms of PSNR, SSIM, and LPIPS.
This indicates that our TALA can significantly enhance the sampling fidelity.

\begin{figure}[htbp]
    \centering

    \begin{minipage}{0.49\linewidth}
        \centering
        \setlength{\abovecaptionskip}{0pt}
        \includegraphics[width=\linewidth]{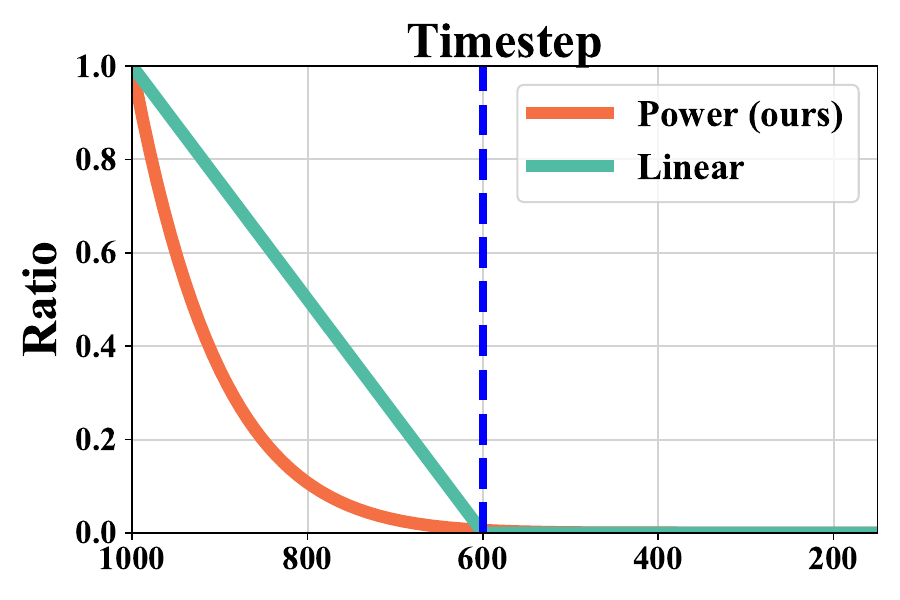}
    \end{minipage}
    % \hfill
    \begin{minipage}{0.49\linewidth}
        \centering
        \setlength{\abovecaptionskip}{0pt}
        \includegraphics[width=\linewidth]{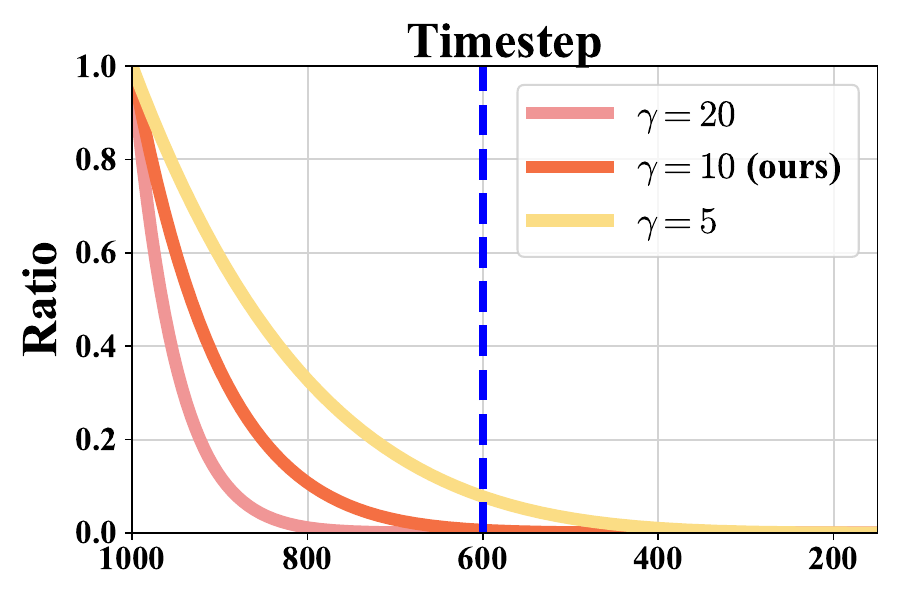}
    \end{minipage}
    
    \begin{minipage}{\linewidth}
    \resizebox{\linewidth}{!}{
        \begin{tabular}{ c | c c | c c}
            \toprule
            \multirow{2}{*}{Methods} & \multicolumn{2}{c|}{Full-reference IQA} & \multicolumn{2}{c}{No-reference IQA}\\
             & PSNR$\uparrow$ & LPIPS$\downarrow$ & NIQE$\downarrow$ & CLIPIQA$\uparrow$\\
            \midrule
            DDPM & 24.77 & 0.3337 & 5.7709 & \blue{0.6881}\\
            TALA(Linear) & 25.24 & 0.3226 & 5.6231 & 0.6647\\
            TALA($\gamma=5$) & 25.16 & \red{0.3197} & 5.7447 & 0.6624\\
            TALA($\gamma=10$) & \red{25.50} & \blue{0.3206} & \red{5.3103} & \red{0.6917}\\
            TALA($\gamma=20$) & \blue{25.38} & 0.3280 & \blue{5.4905} & 0.6811\\
             \bottomrule
        \end{tabular}
    }
    \makeatletter\def\@captype{table}\makeatother
    \caption{Ablation studies on the form and hyper-parameter of the probability function $p(t)$ in our Time-aware Latent Augmentation. The best results are highlighted in red and the second best results are highlighted in blue.}
    \label{tab:TALA}
    \end{minipage}
    \vspace{-10pt}
\end{figure}

Furthermore, we conduct ablation experiments on the time-dependent probability function $p(t)$. We compare our power function against a linear function sharing the same ending point at timestep 600 (as discussed in~\cref{sec:TALA}), yielding $p(t) = max(1-2.5t/T, 0)$.
Then for the empirical parameter $\gamma$, we tested values of 5, 10 (ours), and 20, corresponding to strong, medium, and slight augmentations, respectively. The probability functions and their corresponding results are detailed in~\cref{tab:TALA}.

It can be observed that, compared to the linear function at the same end point (linear), the power function ($\gamma=10$) shows significant advantages in both fidelity and perceptual quality.
Then for the empirical parameter $\gamma$, even with slight augmentation ($\gamma=20$), TALA also significantly improves sampling fidelity. 
The best trade-off between fidelity and visual quality is achieved with $\gamma=10$.
However, when $\gamma$ decreases to 5, its perceptual quality rapidly deteriorates. 
% This is primarily due to the excessive augmentation, which greatly compresses the detail enhancement process in the later timesteps.
This indicates we should confine TALA to the early timesteps.
Excessive augmentation overly emphasizes fidelity, which can adversely affect detail generation in the later timesteps, leading to a decline in perceptual quality.

% In summary, our TALA training strategy effectively bridges the inconsistencies between the diffusion training and the inference process, which greatly increases sampling fidelity while preserving desirable perceptual quality.

\section{Conclusion}
\label{Conclusion}

Pretrained T2I diffusion models offer generative priors for Real-ISR task, yet task inconsistencies hinder existing models from fully exploiting diffusion priors. To bridge the gap between T2I generation and Real-ISR tasks, we present ConsisSR, which adeptly exploit semantic and training-inference consistencies. Utilizing the powerful CLIP image embeddings, our Hybrid Prompt Adapter (HPA) seamlessly integrates both text and image modalities, providing detailed semantic guidance for diffusion process. Furthermore, we introduce Time-Aware Latent Augmentation (TALA) to improve training-inference consistency. By intentionally corrupting the DM inputs in the early timesteps, our model handles diffusion noise as well as accumulated latent, improving sampling fidelity. Our innovative approach achieves state-of-the-art results among existing diffusion models.

{
    \small
    \bibliographystyle{ieeenat_fullname}
    \bibliography{main}
}

\end{document}